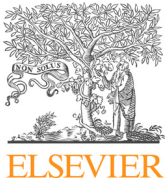
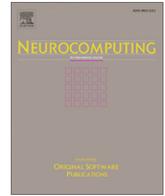

# MultiFace: A generic training mechanism for boosting face recognition performance

Jing Xu [a,1], Tszhang Guo [b,1], Yong Xu [a], Zenglin Xu [a,c,*], Kun Bai [b,*]

[a] *School of Science and Technology, Harbin Institute of Technology, Shenzhen, China*
[b] *Cloud and Smart Industries Group, Tencent, China*
[c] *Artificial Intelligence Center, Peng Cheng Lab, Shenzhen, China*

**ABSTRACT**

Deep Convolutional Neural Networks (DCNNs) and their variants have been widely used in large scale face recognition(FR) recently. Existing methods have achieved good performance on many FR benchmarks. However, most of them suffer from two major problems. First, these methods converge quite slowly since they optimize the loss functions in a high-dimensional and sparse Gaussian Sphere. Second, the high dimensionality of features, despite the powerful descriptive ability, brings difficulty to the optimization, which may lead to a sub-optimal local optimum. To address these problems, we propose a simple yet efficient training mechanism called MultiFace, where we approximate the original high-dimensional features by the ensemble of low-dimensional features. The proposed mechanism is also generic and can be easily applied to many advanced FR models. Moreover, it brings the benefits of good interpretability to FR models via the clustering effect. In detail, the ensemble of these low-dimensional features can capture complementary yet discriminative information, which can increase the intra-class compactness and inter-class separability. Experimental results show that the proposed mechanism can accelerate 2–3 times with the softmax loss and 1.2–1.5 times with Arcface or Cosface, while achieving state-of-the-art performances in several benchmark datasets. Especially, the significant improvements on large-scale datasets(e.g., IJB and MageFace) demonstrate the flexibility of our new training mechanism.

© 2021 Elsevier B.V. All rights reserved.

## 1. Introduction

Large scale face recognition has been an important research topic with emerging and expanded real-world applications. Significant progresses have been made in face recognition with the development of deep Convolutional Neural Networks (CNNs) [1–3]. Unlike the close-set classification task, an open-set task face recognition is more challenging since the number of identities in the training set is limited and far less than that of the real world.

The most commonly used approach for face recognition works in this way—facial images are first encoded into a discriminative feature space, then the recognition results can be obtained according to the similarity among different feature vectors. The key issue of face recognition lies in the feature distribution where we expect the intra-class compactness and inter-class separability properties: features from the same identity are gathered together while those from different identities are scattered away. Since the traditional softmax Loss lacks an adequate power of discrimination, many novel loss functions are proposed to increase the feature discriminative power in the Euclidean margin, including Center loss [3], Contrastive Loss [1] and Triplet loss [2]. Alternatively, novel angular margin-based losses are proposed, such as Sphereface [4], Cosface [5] and Arcface [6]. These loss functions are more stable and robust due to the intrinsic consistency with softmax and they pay more attention to intra-class compactness while maintaining the property of inter-class separation.

However, most of the existing methods lack careful analysis in the selection of spatial dimension. Typically, the dimension of facial feature is set to one single number by empirical study, e.g., 512 in [3,7,5]. Intuitively, low dimension features lack adequate expressive ability but can be easily optimized, while high dimensional features are hard to optimize but can be more powerful to depict faces.

⇑ Corresponding authors at: School of Science and Technology, Harbin Institute of Technology, Shenzhen, China (Zenglin Xu). Cloud and Smart Industries Group, Tencent, China (Kun Bai).
   *E-mail addresses:* zenglin@gmail.com (Z. Xu), kunbai@tencent.com (K. Bai).
 [1] Equal contributions from both authors.

To address this problem of high dimensional features, we propose the MultiFace mechanism that divides the high dimensional space into an ensemble of low dimensional sub-spaces, and then map identities in those sub-spaces respectively, instead of directly mapping in one high dimensional space. We divide the high dimensional feature $x \in \mathbb{R}^d$ into $N$ equal-sized portions, $x = [x^1, x^2, \ldots, x^N]$, each of which sub-features $x^n \in \mathbb{R}^{\frac{d}{N}}$. During training, we add a supervisory loss to each portion to separate different identities, so there are $N$ supervision losses in total. For illustration, we set $N = 4$ in Fig. 1. During inference, as shown in Fig. 2, the similarity between features is decided by a joint decision among sub-groups. We train the CNN under the joint supervision of all the losses and learn all the sub-features from the same neural network backbone. The proposed mechanism can be easily applied to any face recognition models and indeed improves the intepretablity of face recognition models with the clustering effect. In detail, the ensemble of these low-dimensional features can capture complementary yet discriminative information, which reinforces the intra-class compactness and inter-class separability, as verified in Section 4.3.

Experimental results demonstrate that our new models have excellent convergence property, and the collaboration between sub-features improves the face recognition accuracy. We verify the wide applicability of our method which can be combined with different loss functions and diverse CNN backbones. Our contributions are summarized as follows:

- We present a novel training mechanism to promote face recognition performance by using several low dimension spaces to approximate a high dimension space, which is generic and can be applied to many different kinds of neural structures and face recognition losses.
- With our new training methodology, the training of models with the traditional softmax loss can be accelerated 2–3 times, and the training of models with large margin loss such as Arcface and Cosface can be accelerated 1.2–1.5 times.
- We test the proposed method on several widely applicable datasets and achieve the state-of-the-art performance. The significant improvements in large-scale datasets(IJB-B/C and MegaFace1) show the flexibility of the MultiFace mechanism.

## 2. Related work

Recently, face recognition has achieved significant progress thanks to the great success of deep CNN models [8,9]. In DeepFace [10] and DeepID [11], deep CNN models are firstly introduced to learn features on large multi-identities datasets and FR is treated as a multi-class classification problem. The MS-celeb-1 m [12] is the first large scale public training dataset which contains millions of images and thousands of identities. MegaFace [13] is the first large scale public FR benchmark for testing the capacity of models.

Loss functions play an essential role in deep face feature learning. The softmax loss is the most widely used loss function in computer vision community. Given an input feature vector $x_i \in \mathbb{R}^d$ and its corresponding class $y_i$, the softmax loss tries to maximize the posterior probability of the ground-truth class:

$$L_{softmax} = \frac{1}{B}\sum_{i=1}^{B} -\log p_i = \frac{1}{B}\sum_{i=1}^{B} -\log \frac{e^{W_{y_i}^T x_i + b_{y_i}}}{\sum_{j=1}^{C} e^{W_j^T x_i + b_j}}, \quad (1)$$

where $B$ denotes the batch size and $C$ denotes the number of classes. $W_j \in \mathbb{R}^d$, ($j \in \{1, 2, \ldots, C\}$) denotes the $j^{th}$ column of the classifier $W \in \mathbb{R}^{C \times d}$ and $b_j \in \mathbb{R}^1$, is the bias corresponding to class $j$. Traditional softmax loss can well solve close-set classification problems, but it is insufficient to explore the full discriminating information between features. Hence, several softmax-free loss functions are proposed such as the Contrastive loss [1] and the Triplet loss [2] where they use metric learning for directly optimizing the feature space. They minimize the distance of two features from positive pairs while maximizing the distance from negative pairs. Since the amount of negative pairs is naturally larger than positive pairs, these methods are at high risk of over-fitting and hard to optimize. Center loss [3] proposes an online-sampling method for solving this problem and integrates the softmax loss to strengthen the inter-class separation. However, the online sampling pipeline is time-consuming and hard to balance the scales of supervisory signals between the softmax and the metric learning loss. CDA [14] and the Git loss [15] are other methods combining the center loss and the softmax loss which encourages the intra-class compactness and inter-class separability. It is worth noting that our method is compatible with these methods mentioned above. The main contribution of our mechanism is that we divide high-dimensional feature into sub-spaces, which absorbs both the advantages of high-dimensional and low-dimensional optimization space. We can further improve the capability of our model by altering the loss functions which supervise the low-dimensional features. We will explore this in future work.

Alternatively, the angular-based softmax loss assumes that each person can be represented by a corresponding center vector and the distances between negative pairs can be simplified as the angle between different center vectors which can be easily acquired. Furthermore, they design several manual margins to learn more discriminate decision boundary. The SphereFace [4] is the first method which proposes a multiplicative angular margin penalty to strengthen extra intra-class compactness and inter-class discrepancy simultaneously. Cosface [5] directly adds cosine margin penalty to the target logit and is much easier to train than SphereFace. Arcface [6] proposes an additive angular margin and releases a cleaned version of the MS-celeb-1 m which is quite beneficial for face recognition community. Following the basic concept of Arcface and Cosface, several modified versions have been proposed like the AdaCos [16], AdaptiveFace [17].

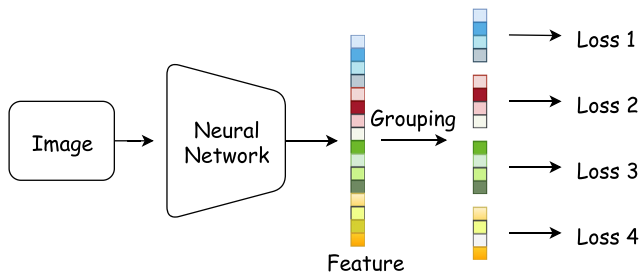

**Fig. 1.** An illustration of the proposed training mechanism. The main idea of MultiFace is that instead of directly learning a high-dimensional feature vector supervised by a loss function, we equally group the high-dimensional vector into 4 low dimension feature supervised by 4 loss functions separately.

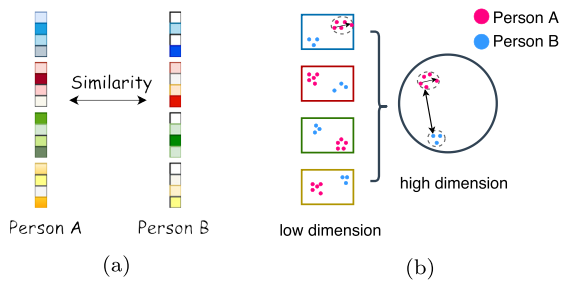

**Fig. 2.** (a) demonstrates that the similarity between two vectors is jointly decided by the sub-features. (b) shows that low dimension sub-space is better at capturing the intra-class compactness of the encoded features while the inter-class separation property can be maintained in high dimension.

# 3. Our method

In this section, we revisit the state-of-the-art large margin-based loss functions and propose our simple yet effective method to extend a new training mechanism on these promising functions.

## 3.1. MultiFace large margin loss functions

Traditional softmax loss emphasizes inter-class separation on close-set classification, which is inadequate for open-set face recognition. To guarantee small intra-class and large inter-class distance properties in face recognition, researchers propose large angular margin loss functions that modify the basic formula of softmax loss by adding some extra constraints.

In large margin loss, the bias terms are usually set to zero, the class center $W_j$ and face feature $x_i$ are always uniformly normalized by their L2-norm values. Therefore the inner product of $W_j^T x_i$ can be simplified as:

$$W_j^T x_i = |W_j| \cdot |x_i| \cdot \cos(\theta_{j,i}) = \cos(\theta_{j,i}), \quad (2)$$

where $\theta_{j,i}$ denotes the angle between $W_j$ and $x_i$. Large margin losses reformulate $\cos(\theta_{j,i})$ with some manual prior knowledge:

$$\cos(\theta_{y_i,i}) \rightarrow s[\cos(m_1 \theta_{y_i,i} + m_2) + m_3] \quad (3)$$
$$\cos(\theta_{j,i}) \rightarrow s \cos(\theta_{j,i}), j \neq y_i, \quad (4)$$

where $s, m_1, m_2, m_3$ are four hyper-parameters for balancing the difficulty between intra-class compactness and learning stability. The final formula of large margin loss(shorted as LML) is:

$$L_{LML} = \frac{1}{B} \sum_{i=1}^{B} -\log p_i \quad (5)$$

$$p_i = \frac{e^{s[\cos(m_1 \theta_{y_i,i} + m_2) + m_3]}}{e^{s[\cos(m_1 \theta_{y_i,i} + m_2) + m_3]} + \sum_{j=1, j \neq y_i}^{C} e^{s \cos(\theta_{j,i})}}.$$

Specifically, SphereFace [4] fixes $m_2 = m_3 = 0$, Cosface [5] fixes $m_1 = 1, m_2 = 0$ and Arcface [6] fixes $m_1 = 1, m_3 = 0$.

Although large margin loss functions have achieved significant progress in face recognition recently, it is a troublesome optimization problem that directly maps a face image to a hyper-sphere manifold, especially in a high-dimensional space. As we mentioned above, the typical setting of embedding size for face feature is 512-dimension, which is quite hard to optimize at the beginning of training. To tackle this problem, we equally divide the $d$-dimension feature vector $x_i$ into $N$ groups:

$$x_i = [x_i^1, x_i^2, \ldots, x_i^N], \quad (6)$$

where $x_i^n \in \mathbb{R}^{\frac{d}{N}}, n \in \{1, 2, .., N\}$.

Correspondingly, we use $N$ supervisory loss functions with $N$ weight matrix to learn the vectors $[x_i^1, x_i^2, \ldots, x_i^N]$ individually. In our MultiFace large margin loss (shorted as MLML) function, we define $N$ small weight matrices($W^n \in \mathbb{R}^{C \times \frac{d}{N}}$) and each matrix focuses on learning its own discriminated sub-vector $x_i^n$ by the corresponding supervisory loss. The whole training loss can be formulated as:

$$L_{MLML} = \sum_{n=1}^{N} L_{LML}^n, \quad L_{LML}^n = \frac{1}{B} \sum_{i=1}^{B} -\log p_i^n, \quad (7)$$

$$p_i^n = \frac{e^{s[\cos(m_1 \theta_{y_i,i}^n + m_2) + m_3]}}{e^{s[\cos(m_1 \theta_{y_i,i}^n + m_2) + m_3]} + \sum_{j=1, j \neq y_i}^{C} e^{s \cos(\theta_{j,i}^n)}}.$$

As we know, the new training mechanism is easy to extend to other Euclidean margin based loss functions, like the Center loss, the Triplet loss and so on. During inference, the similarity between $x_i$ and $x_j$ is calculated by the following formula:

$$x_i x_j = \sum_{n=1}^{N} x_i^n x_j^n. \quad (8)$$

## 3.2. Analysis on feature dimension selection

Given $m$ training persons, both the angular-based loss and margin-based loss attempt to distribute their feature vectors as uniformly as possible in a $n$-dimension sphere, which is easy to satisfy when $n$ is large enough. However, it increases the risk of over-fitting. To select a good trade-off $n$, we propose a new problem: how many points at most could be located on the surface of an $n$-$d$ sphere such that the minimum angle between any two points is lower than a fixed threshold $\theta$ such as $\frac{\pi}{3}$ in Fig. 3.

The surface area of an uniform $n$-$d$ sphere $S_n$ is:

$$S_n = \frac{2\pi^{\frac{n}{2}}}{\Gamma(\frac{n}{2})}, \quad (9)$$

where the $\Gamma(\cdot)$ denotes the gamma function.

Given an fixed point $\alpha$ in $S_n$ and a subset $t$ in $S_n$ where $t$ satisfy:

$$\forall \beta \in t, (\beta - \alpha)^2 \leqslant \left(2 \sin(\frac{\theta}{4})\right)^2. \quad (10)$$

The area of $t$ is:

$$S_n^t = S_{n-1} \int_{\cos(\frac{\theta}{2})}^{1} (1-x^2)^{\frac{n-2}{2}} dx. \quad (11)$$

For more details please refer to Chapter 1.2.5 in the notes given by Professor Venkatesan Guruswami and Ravi Kannan.[2]

As shown in Fig. 3, the problem can be approximated as "how many $t$ can be distributed in $S_n$". And the maximum number $m^*$ can be approximated as:

$$m^* \approx \frac{S_n}{S_{n-1}} \frac{1}{\int_{\cos(\frac{\theta}{2})}^{1} (1-x^2)^{\frac{n-2}{2}} dx}. \quad (12)$$

For $n = 128$ and $\theta = \frac{\pi}{3}$ then $m^* \approx 10^{22}$. Therefore we can easily satisfy the inter-class separability with 128 dimensions and strengthen the intra-class compactness by assembling these sub-group features trained with multi-supervised signal.

# 4. Experiments

In this section, we conduct extensive experiments on several widely-used face recognition datasets with different CNN architectures to demonstrate the broad applicability of our proposed method. Our codes will be publicly available.

## 4.1. Implementation details

### 4.1.1. CNN architecture

The expressive power of the network directly affects the performance of face recognition. To show the extensive applicability of our method, we illustrate the experiment results with backbones of different sizes. For lightweight CNN network, we choose Mobile-FaceNet [18] with 1.2 M parameters. For the large-scale network, we choose the widely used CNN architectures ResNet 100 same

---

[2] https://www.cs.cmu.edu/venkatg/teaching/CStheory-infoage/chap1-high-dim-space.pdf.

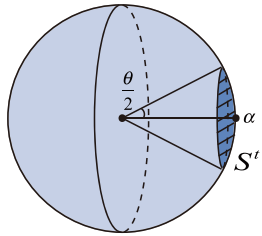

**Fig. 3.** The area of $S^t$ in 3-*d* for illustration.

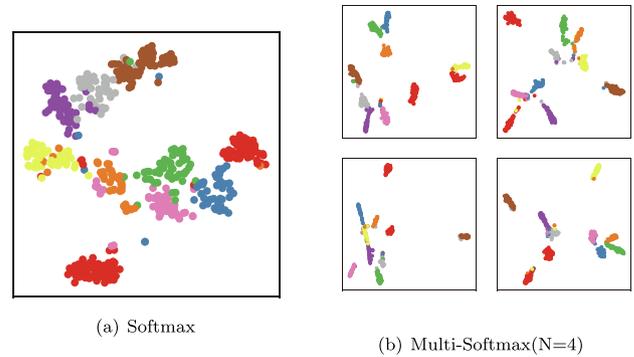

(a) Softmax          (b) Multi-Softmax(N=4)

**Fig. 4.** Comparison of the features distribution on MNIST between baseline softmax loss and our Multi-Softmax(N = 4). All of the embedding features are reduced to 2-*d* using t-SNE for visualization. 4a shows the distribution of 32-*d* features supervised by a vanilla softmax loss, 4b shows the distribution of sub-features which supervised by Mul.ti-Softmax loss.

in Arcface [6] with 65 M parameters.

*4.1.2. Datasets*

We train our models on MS1MV2 [6] dataset which is a clean version of MS-Celeb-1 M refined by the author of Arcface [6] with 5.8 M images from 85 K individuals. During evaluation, we use the most widely used face verification dataset LFW [19], CFP-FP [20], AgeDB-30 [21], to show the improvement after combining our method with other loss functions. The most widely used LFW verification dataset includes 13 K images from 5.7 K identities. The CFP-FP contains 7 K images from 500 identities and the AgeDB-30 contains 16 K images from 568 identities. Extensively, we evaluate our models on the CPLFW [22], CALFW [23] and large-scale dataset MegaFace [24], IJB-B/C [25] to demonstrate the generalization ability of our method.

*4.1.3. Experimental setting*

Input face images in both training and testing are detected by MTCNN [26] and aligned through 5 landmarks. Then the input images are cropped into 112×112 and each pixel is normalized by subtracting 127.5 and dividing by 128. The outputs of all the CNN networks are 512-*d* embedding features. The margin is set to 0.3 in our Multi-Cosface and Multi-Arcface and we keep the dropout rate to 0.4. We use SGD algorithm with momentum 0.9 and set the weight decay to 4e-5. The batch size is 180 and the learning rate starts from 0.05 and is divided by 10 at 320 K, 480 K and 580 K iterations. All the experiments are implemented in Pytorch platform.

*4.2. Analysis on MNIST*

We conduct an experiment on MNIST [27] to demonstrate the superiority of our proposed method. We adopt the lightweight LeNet [28] as the mapping network. We map the images to 32-*d* features using original softmax and our Multi-Softmax(N = 4) separately.

We compare the distribution of embedding features of different digit classes in Fig. 4. We constrain all the features to a 2-*d* space using t-SNE for ease of visualization. As shown in Fig. 4a, the margins between different classes are not clear enough with original softmax loss. In Fig. 4b, after dividing features into 4 groups, the digits belong to the same class are gathered more compactly in each sub-group. Meanwhile, the 4 sub-groups jointly determine the final classification result. In the left upper of Fig. 4b, some digits belong to a purple class are close to the grey class. At the same time, in another subgroup, the upper right of Fig. 4b, the purple class, and grey class are separated by a great distance.

*4.3. Analysis on feature distribution*

In this section, we explore the distribution of learned features with different embedding dimensions and demonstrate the superiority feature distribution learned by our MultiFace mechanism. Without loss of generality, the softmax loss is used for illustration with MobileFaceNet as the embedding network. To see the feature distribution more intuitively, we randomly select 0.3 M positive pairs (two images from a same person) and negative pairs (two images from different persons) separately from training dataset and draw the angle(the distance between each pair) distributions in Fig. 5.

As we know, inter-class separation and intra-class compactness both decide the recognition performance. High-dimensional feature has adequate space to separate numerous people, but learning the compact representation in high dimension for each person is a hard optimization problem especially for lightweight networks. Lower dimension space is easier to learn the intra-class compactness, but increasing the risk of over-fitting. As we stated in Section 3.2, the inter-class dispersion can be maintained with a relative low dimension embedding space, e.g., 128-*d*. As shown in Fig. 5, the 128-*d* feature with the red line has lower mean of angles of positive pairs while somewhat maintains the distances between negative pairs compared with the 512-*d* with the blue line.

By dividing the high dimension space into several low dimensions with multiple supervised signals, our MultiFace inherits both their respective advantages. Shown in Fig. 5, MultiFace, which divides the 512-*d* into 4 128-*d* sub-features with the green line, learns more compact distances between positive pairs, and still holds the dispersion of negative pairs. That shows MultiFace can notably enhance the discriminative power of learned features. The performance boosting is derived from:

(1) The independence of all supervised losses: the sub-group features are potentially independent which are beneficial for feature boosting. We measure the correlation of groups of features and find that they are almost orthogonal between each other. During the training procedure, the backbone structure should not only optimize the ability of facial representation but also maintain the independence among sub-group features. (2) Data augmentation: Each batch of training data generates four times magnitude of independent gradients which makes the training process more stable and reduces the risk of over-fitting. Therefore MultiFace can strengthen the model's generation ability and lead to a better performance.

*4.4. Analysis on training process*

In this section, we want to figure out how multi-signals cooperate with each other and speed up training. Fig. 6a shows the gradient curves during training in MultiFace (N = 4) with softmax loss.

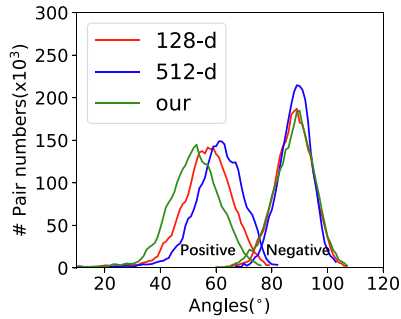

**Fig. 5.** Angle distributions of random positive pairs and negative pairs (0.3 M) from MS1MV2, all angles are represented in degree. We compare the angle distributions of 128-*d*, 512-*d* and our MultiFace(4-128d).

Each sub-feature is supervised by a head matrix classifier. We average the absolute value of the weight gradients in each head, denoted by head$_i$, $i \in \{1, 2, 3, 4\}$.

As we can see, at the beginning of the training, the gradient in one head is higher and the others are relative lower. That means the model focuses on learning one of the sub-features first and the optimization of the other heads is somewhat suppressed. Encoding images into a low dimension space is much easier and faster. But after a few steps, when reaching a balance between one head and the backbone, the gradients of the other are gradually tending to be the same as the first one: all the heads are updated almost simultaneously. The collaboration speeds up the training process and makes the training process more stable and faster.

We draw the validation accuracy curves and compare our method with traditional softmax, Cosface and Arcface. Intuitively, with the help of our method, softmax loss can be accelerated 2–3 times for convergence in Fig. 6b, Arcface and Cosface can be accelerated 1.2–1.5 times in Fig. 6c. During inference, the supervised classifiers are remobed. Our feature extract backbones are the same as the original methods, so the computational time is equal with baseline models.

### 4.4.1. Ablation study on number of groups

We explore the setting of the number of groups with MobileFaceNet in Table 1. With the help of MultiFace training mechanism, Multi-Softmax(N = 4) increases accuracy on LFW by 0.25%, CFP-FP by 2.67% and AgeDB-30 by 1.62%. In face verification task, we want to separate positive pairs from the negative ones according to the distance between each pair and the verification error comes from the intersection of the positive and the negative pairs. We show the angle(the distance between each pair) distributions with different groups(N) on LFW(the first line), CFP-FP(the second line) and AgeDB-30(the third line) in Fig. 7. With the increasing of N, the intersection parts in all three datasets decrease first and increase

**Table 1**
The ablation study on different number of groups(N) and different losses with MobileFaceNet.

| Loss Function | LFW | CFP-FP | AgeDB-30 |
| --- | --- | --- | --- |
| Softmax | 99.18 | 92.80 | 93.70 |
| Multi-Softmax (N = 2) | 99.43 | 94.09 | 95.18 |
| Multi-Softmax (N = 4) | 99.43 | **95.47** | 95.32 |
| Multi-Softmax (N = 8) | **99.52** | 95.34 | **95.52** |
| Multi-Softmax (N = 16) | 99.35 | 93.99 | 94.10 |
| Cosface | 99.43 | 92.71 | 96.03 |
| Multi-Cosface (N = 4) | **99.62** | **94.09** | **96.12** |
| Arcface | 99.47 | 92.47 | 96.05 |
| Multi-Arcface (N = 4) | **99.60** | **93.36** | **96.38** |

later and the verification accuracies increase first and decrease later as shown in Table 1.

The number of groups affects verification accuracies. As we mentioned in Section 4.3, lower dimension space with larger N is easier to learn intra-class compactness, but increase the risk of over-fitting. Besides, the setting of the hyper-parameter *N* also highly correlates to the expressive ability of the backbone network. So, we choose N = 4 for small backbone like MobileFaceNet, and N = 2 for larger backbone like ResNet100.

### 4.4.2. Ablation study on losses

Cosface [5] and Arcface [6] are two margin-based losses that have achieved outstanding performance recently. As shown in Table 1 our methods improve the performance of both Cosface and Arcface with faster convergence speed, which shows the extensive generalization capability.

### 4.5. Evaluation results

#### 4.5.1. Results on CALFW and CPLFW

We report the results on LFW [19], CALFW [23] and CPLFW [22] comparing with other state-of-the-art methods. The face verification datasets CALFW and CPLFW have the same identities as LFW. The CALFW contains 12.3 K face images with higher pose variations, and The CPLFW contains 11.7 K images with higher age variations. As shown in Table 2 and 3, our Multi-Arcface (N = 2) with ResNet100 (shorted as R100) achieves the accuracy of 99.83% on LFW and boosts the verification accuracy to 96.08% on CALFW and 93.10% on CPLFW.

#### 4.5.2. Results on MegaFace Challenge1

MegaFace Challenge1 [24] is a challenging benchmark dataset for large-scale face identification and verification. It contains a gallery set and a probe set. There are 1 M images of 690 K different individuals in the gallery set, and in the probe set, there are 107 K face images of 530 identities from Facescrub dataset. In the face identification task, we compute the similarity between the face image in the probe dataset with each face image in the gallery. In the face verification task, we consider whether a pair of face images belong to the same identity.

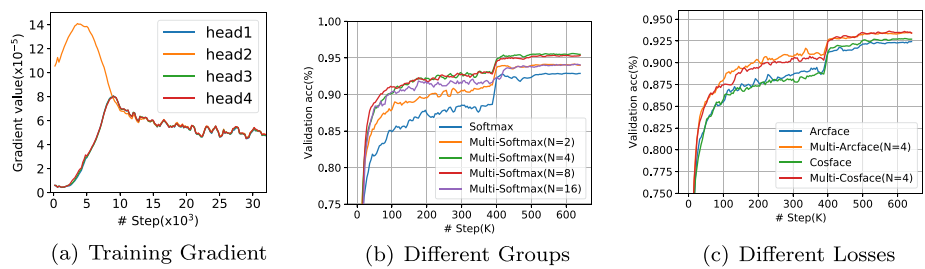

(a) Training Gradient    (b) Different Groups    (c) Different Losses

**Fig. 6.** (a) The gradient curves during training in MultiFace(N = 4) with softmax loss. (b) The curves of verification results on CFP-FP with vanilla softmax in different setting N. (c) The curves of verification results on CFP-FP with Multi-Arcface(N = 4) and Multi-Cosface(N = 4) compared with their original ones.

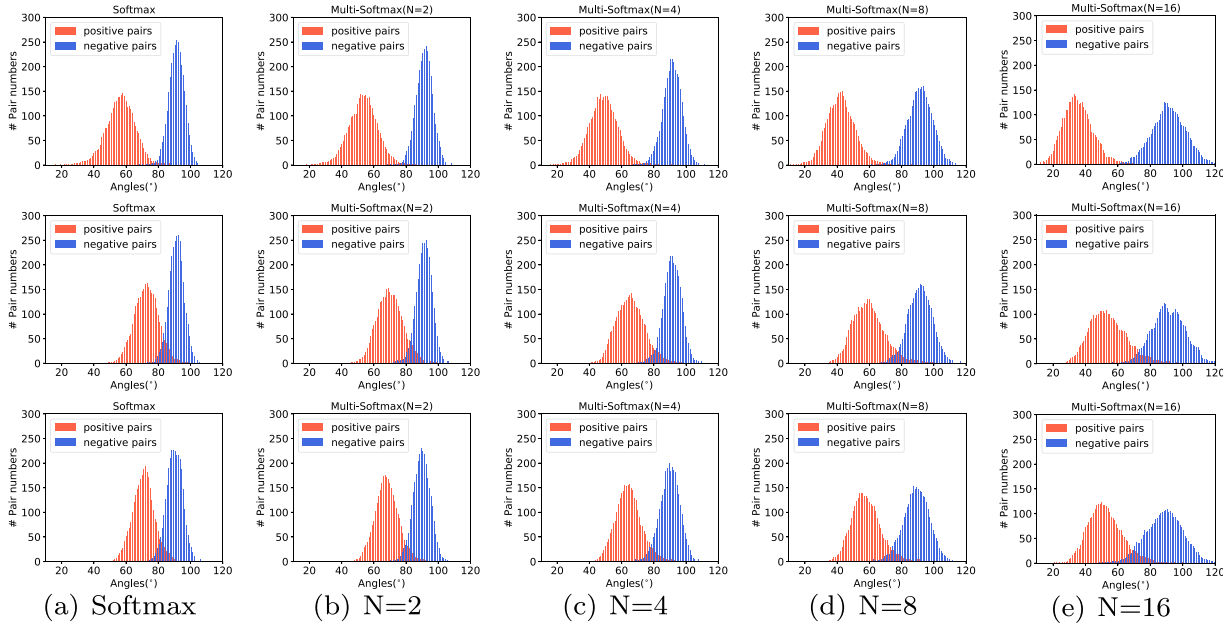

**Fig. 7.** The angle distributions with different groups N on LFW(the first line), CFP(the second line) and AgeDb-30(the third line). With the increasing of N, the intersection parts between positive and negative pairs in all three datasets decrease first and increase later. Correspondingly, the verification accuracies increase first and decrease later as shown in Tab.le 1.

**Table 2**
Verification performance of state-of-the-art face recognition models on LFW.

| Method | #Image | LFW |
|---|---|---|
| FaceNet [2] | 200 M | 99.63 |
| Center Loss [3] | 0.7 M | 99.28 |
| Range Loss [7] | 5 M | 99.52 |
| Marginal Loss [29] | 3.8 M | 99.48 |
| Long Tail [30] | 0.55 M | 99.48 |
| RegularFace [31] | 3.3 M | 99.61 |
| AdaptiveFace [17] | 5 M | 99.62 |
| AdaCos [16] | 0.5 M | 99.71 |
| UniformFace [32] | 6.1 M | 99.80 |
| Cosface [5] | 5 M | 99.73 |
| Arcface [6] | 5.8 M | **99.83** |
| Multi-Arcface | 5.8 M | **99.83** |
| Multi-Cosface | 5.8 M | 99.77 |

**Table 3**
Verification performance of state-of-the-art face recognition models on CALFW and CPLFW.

| Method | CALFW | CPLFW |
|---|---|---|
| Center Loss [3] | 85.48 | 77.48 |
| SphereFace [4] | 90.30 | 81.40 |
| VGGFace2 [33] | 90.57 | 84.00 |
| Arcface [6] | 95.45 | 92.08 |
| Multi-Arcface | **96.08** | **93.32** |
| Multi-Cosface | 96.07 | 93.00 |

**Table 4**
Face identification and verification evaluation under large protocal(more than 0.5 M images in training dataset) of different methods on MegaFace Challenge1 using FaceScrub as the probe set. "Id" refers to the rank-1 face identification accuracy with 1 M distractors, and "Ver" refers to the face verification TAR at $10^{-6}$ FAR.

| | Method (Large protocol) | Id(%) | Ver(%) |
|---|---|---|---|
| Not Refined | FaceNet [2] | 70.49 | 86.47 |
| | SphereFace [4] | 76.65 | 92.32 |
| | UniformFace [32] | 79.98 | 95.36 |
| | Cosface [5] | **82.72** | 96.56 |
| | Arcface [6], R100 | 81.03 | 96.98 |
| | Multi-Arcface | 80.93 | 97.16 |
| | Multi-Cosface | 80.91 | **97.37** |
| Refined | AdaptiveFace [17] | 95.02 | 95.61 |
| | AdaCos [16] | 97.41 | – |
| | Cosface [5], R100 | 97.91 | 97.91 |
| | Arcface [6], R100 | 98.35 | 98.48 |
| | Multi-Arcface | 98.44 | 98.72 |
| | Multi-Cosface | **98.62** | **98.85** |

In Table 4, we compare the identification result Rank-1 (1 M distractors) and the verification result TAR (@FAR=$10^{-6}$) of Multi-Arcface (N = 2) and Multi-Cosface (N = 2) with the previous state-of-the-art models. Since there are many face images with wrong labels in the original MegaFace dataset, a refined version of MegaFace is released by Arcface [6] authors. On the original MegaFace dataset with wrong labels, which is not be refined, Multi-Cosface and Multi-Arcface outperform the original Cosface and Arcface on verification task. On the refined version of MegaFace, our models boost the performance on both identification and verification tasks. Compared with Arcface, using the same network(ResNet100) and training dataset, Multi-Arcface further improves the Rank-1 from 98.35% to 98.44% and TAR from 98.48% to 98.72%. Meanwhile, the Multi-Cosface boosts the Rank-1 to 98.62% and TAR to 98.85%, setting a new baseline.

*4.5.3. Results on IJB*

We also evaluate Multi-Arcface and Multi-Cosface on large-scale dataset IJB-B/C. The IJB-B dataset has 1.8 K subjects with 21.8 K images and 55 K frames from 7.0 K videos. IJB-C, the extension of IJB-B, contains 3.5 K subjects with 31.3 K images and 117.5 K frames from 11.8 K videos. Shown in Table 5, our Multi-Arcface (N = 2) and Multi-Cosface (N = 2) improve the verification results TAR(@FAR=$10^{-4}$) of original Arcface and Cosface. Compared with Arcface, with the same CNN backbone(ResNet100) and training dataset, Multi-Arcface gets 0.93% improvement on IJB-B and 0.78% on IJB-C. At the same time, it achieves the state-of-the-art performance on both IJB-B (95.13%) and IJB-C (96.38%).

**Table 5**
The verification results on IJB-B and IJB-C. The TAR is reported with $10^{-4}$ FAR.

| Method | IJB-B | IJB-C |
| --- | --- | --- |
| VGGFace2 [33] | 80.00 | – |
| FPN [34] | 83.02 | – |
| ResNet50 [35] | 83.10 | 86.20 |
| SENet50 [36] | 84.90 | 88.50 |
| PFN [37] | 84.50 | – |
| AdaCos [16] | – | 92.40 |
| Arcface [6] | 94.20 | 95.60 |
| Multi-Arcface | **95.13** | **96.38** |
| Multi-Cosface | 94.94 | 96.17 |

## 5. Conclusion

In this paper, we propose a simple yet effective training mechanism called MultiFace, where we approximate the original high-dimension space by several low-dimension dense spaces. Our model is beneficial to obtain better intra-class compactness and inter-class separation. Ablation studies figure out that our method is generic and can be applied to many different kinds of neural structures and face recognition losses. Combined with our method, several state-of-the-art face recognition methods such as Cosface and Arcface can achieve higher accuracy in several benchmarks and converge faster at the same time, e.g. 2–3 times for softmax and 1.2–1.5 times for Arcface or Cosface.

**CRediT authorship contribution statement**

**Jing Xu:** Conceptualization, Methodology, Software. **Tszhang Guo:** Conceptualization, Methodology, Software. **Yong Xu:** review and editing. **Zenglin Xu:** Supervision, Writing - review & editing. **Kun Bai:** Supervision, Resources.

**Declaration of Competing Interest**

The authors declare that they have no known competing financial interests or personal relationships that could have appeared to influence the work reported in this paper.

**Acknowledgement**

This work was partially supported by the National Key Research and Development Program of China (No. 2018AAA0100204), and a fundamental Program of Shenzhen Science and Technology Innovation Commission (No. ZX20210035).